%% file: main.tex
\documentclass[10pt]{article}

\usepackage[letterpaper,textwidth=5.5in,textheight=9in,centering]{geometry}
\usepackage{times}
\usepackage{graphicx}
\usepackage{xcolor}
\input{math_commands.tex}

\usepackage{natbib}
\usepackage{url}
\usepackage{booktabs}
\usepackage{placeins}
\usepackage{hyperref}
\hypersetup{colorlinks=true,linkcolor=blue,citecolor=blue,urlcolor=blue,
  pdftitle={R2-WAM: Repair-and-Reject Post-Training for World Action Models},
  pdfauthor={Ruiyan Xu, Haisheng Su, Sixu Lin, Zhaokun Yue, Chengming Hu, Xin Jin, Guiliang Liu}}

\newcommand{\method}{\ensuremath{R^2}\mbox{-WAM}}

\title{\method: Repair-and-Reject Post-Training for World Action Models}
\author{%
Ruiyan Xu\textsuperscript{1}, Haisheng Su\textsuperscript{2},
Sixu Lin\textsuperscript{1}, Zhaokun Yue\textsuperscript{3}\\
Chengming Hu\textsuperscript{2}, Xin Jin\textsuperscript{4},
Guiliang Liu\textsuperscript{1}\thanks{Corresponding author: \href{mailto:liuguiliang@cuhk.edu.cn}{\texttt{liuguiliang@cuhk.edu.cn}}.}\\[0.6em]
\small\textsuperscript{1}The Chinese University of Hong Kong, Shenzhen\\
\small\textsuperscript{2}Manifold AI\\
\small\textsuperscript{3}Southeast University\\
\small\textsuperscript{4}Chang'an University
}
\date{}

\begin{document}
\maketitle

\begin{abstract}
World Action Models (WAMs) emerge as a promising foundation for policy refinement by predicting the consequences of sampled actions.
However, visually plausible predictions can mislead policy refinement if they fail to reflect the input actions.
To address this mismatch, we introduce \method{}, a two-stage repair-and-reject post-training framework that first improves the consistency of predicted futures with input actions, then uses these futures to select inferior action samples for negative fine-tuning.
The repair stage grounds imagination in observed robot behavior through a kinematic alignment score that measures agreement between predicted and demonstrated motion, enabling the predicted video to faithfully reflect its input actions.
Using the repaired video model, the rejection stage compares imagined outcomes of sampled and demonstrated actions, selectively applying negative fine-tuning to samples whose predicted task progress falls below the demonstrated reference by a prescribed margin.
Together, the two stages extend video prediction from representation learning to consequence-based policy refinement without additional environment interaction or changes to the inference procedure.
\method{} achieves 93.8\% average success on RoboTwin~2.0
across clean and randomized settings.
On the long-horizon real-world Fold Shirt task, it achieves
87.5\% average success, compared with 0\% for Fast-WAM.
Our project page is available at \href{https://r2-wam.github.io/}{R2WAM}.
\end{abstract}

\input{sections/introduction}
\begin{figure}[!ht]
    \centering
    \includegraphics[width=0.99\linewidth]{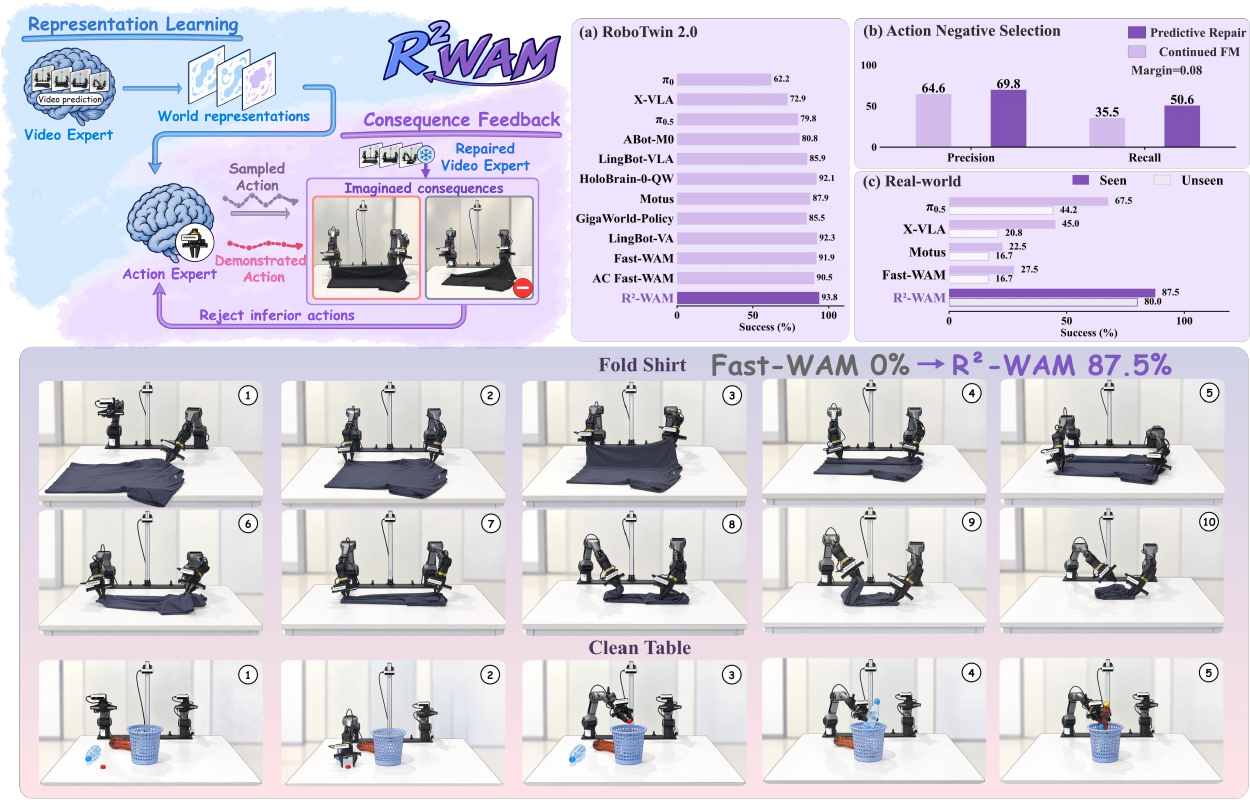}
    \caption{\method{} extends video prediction from representation
learning to consequence feedback for policy post-training.
(a) Average success on RoboTwin~2.0.
(b) Action negative-selection precision and recall against
execution-based STEAM labels.
(c) Real-world success under seen and unseen conditions.
Bottom: execution sequences on Fold Shirt and Clean Table.}
    \label{fig:teaser}
\end{figure}
\input{sections/related_work}
\input{sections/preliminary}
\input{sections/method}
\input{sections/experiments}
\input{sections/conclusion}

\subsection*{AI Use Statement}
We used generative AI tools to improve the clarity, grammar,
and readability of the manuscript.
The authors reviewed and revised the AI-assisted text and
take full responsibility for the final manuscript.

\bibliography{iclr2027_conference}
\bibliographystyle{iclr2027_conference}

\appendix
\input{sections/method_appendix}
\input{sections/experiments_appendix}

\end{document}

%% file: math_commands.tex
\usepackage{amsmath,amsfonts,bm}

\def\eqref#1{equation~\ref{#1}}

\def\1{\bm{1}}

\def\mK{{\bm{K}}}

\DeclareMathAlphabet{\mathsfit}{\encodingdefault}{\sfdefault}{m}{sl}
\SetMathAlphabet{\mathsfit}{bold}{\encodingdefault}{\sfdefault}{bx}{n}



%% file: sections/introduction.tex
\section{Introduction}
\label{sec:introduction}

World Action Models (WAMs) have recently emerged as a promising approach to robotic control by guiding action generation with predicted visual outcomes.
Existing WAMs typically learn to predict video and actions jointly in pixel or latent space \citep{ye2026world,zhu2025unified,Bi_2026_CVPR,li2026causal,yuan2026fast}, thereby grounding action generation in predicted scene dynamics.
Beyond direct robot control, recent studies have explored using action-conditioned world models as simulators to predict the consequences of candidate actions and provide feedback for policy improvement \citep{guo2026ctrl,yang2026rise}.
This is particularly useful when demonstrations provide limited coverage of execution errors and collecting physical trials is costly
\citep{chandra2025diwa,jiang2025world4rl,xiao2025worldenv,jiang2026wovr}.

A key prerequisite for effective policy post-training with a world model is that it faithfully predicts how the environment changes in response to actions.
Although modern video generation models can produce visually plausible predictions~\citep{wan2025}, they do not necessarily preserve physical properties such as kinematics and dynamics.
Consequently, generated videos may appear realistic while failing to reflect the supplied actions \citep{yang2026mirabench}.
Evaluating actions with such predictions can mistake video model errors for policy failures, causing the policy to learn from outcomes its actions would not produce\citep{jiang2026wovr,chen2026worldsync}.

To reduce mismatch between imagined and actual consequences, prior methods ground generated videos in robot motion using inverse dynamics \citep{wang2026eva} or extracted 3D flow \citep{dharmarajan2025dream2flow}.
These methods rely on auxiliary components to obtain motion feedback, but do not ensure that the video faithfully follows a supplied action.
Recent methods improve imagined feedback by fine-tuning world models on real policy rollouts \citep{guo2026vlaw}, updating the model as the policy evolves from task-critical keyframes \citep{jiang2026wovr}, or training on diverse actions with supervision~\citep{chen2026worldsync}.
These methods improve action grounding, but leave open how observed demonstration outcomes can identify unreliable predictions before they are used to evaluate sampled policy actions.

To fill this gap, we propose repairing the video expert using observed consequences of demonstrated actions, then using the repaired expert to reject sampled actions with inferior predicted task progress (Figure~\ref{fig:teaser}).
To this end, we introduce \method{}, a two-stage repair-and-reject post-training framework comprising predictive repair and action rejection (Figure~\ref{fig:pipeline}), comprising the following modules:

1) {\it Predictive repair} turns each demonstration into a test of action fidelity: given the same observation and action, the video expert generates a future that can be compared with what the robot actually did. 
A kinematic alignment score measures whether the generated robot motion follows the observed motion and selects predictions with substantial discrepancies for negative fine-tuning.
This stage improves the video expert before its imagined consequences are used to judge sampled policy actions, reducing the risk that prediction errors become misleading training signals.

2) {\it Action rejection} uses the repaired video expert to judge what sampled actions are likely to achieve. 
With the expert frozen, we predict futures for sampled and demonstrated actions from the same initial observation, then compare their task progress using STEAM~\citep{liu2026steam}. 
Sampled actions whose predicted progress falls below the demonstrated reference by a prescribed margin are selected for negative fine-tuning, while demonstrations remain the positive targets. 
This directs policy updates toward actions with evidence of inferior consequences.

Together, the two stages turn video prediction into feedback for
policy post-training without additional environment interaction or
changes to the inference procedure.
We conduct an extensive empirical study across 50 RoboTwin~2.0 tasks
and two real-world manipulation tasks, evaluating action fidelity,
negative-selection accuracy, and downstream policy success.
\method{} improves average success over action-conditioned Fast-WAM
by 3.3 percentage points in simulation and raises average success
on the long-horizon real-world Fold Shirt task from 0\% with
Fast-WAM to 87.5\%.

%% file: sections/related_work.tex
\section{Related Work}
\label{sec:related_work}

\paragraph{World action models.}
World action models couple future prediction with action generation
through joint diffusion modeling of video and
actions~\citep{ye2026world,zhu2025unified},
causal modeling~\citep{li2026causal}, and latent action
representations~\citep{Bi_2026_CVPR}. WorldScape Policy
2.0~\citep{su2026worldscape} extends this framework with memory and multimodal
conditioning. Predictions can also guide control through visual
representations~\citep{hu2024video} or explicit motion plans:
RoboFlow4D~\citep{lin2026roboflow4d} predicts multi-frame 3D motion to condition
action generation. Fast-WAM~\citep{yuan2026fast} couples video and action
experts through shared attention, using video prediction to learn world
representations without generating future videos at deployment.
We build on Fast-WAM to use predicted consequences
for policy post-training.

\paragraph{Policy post-training.}
RECAP~\citep{intelligence2025pi} refines policies through
advantage-conditioned learning from experience.
World-model-based approaches use synthetic rollouts and predicted
consequences for policy fine-tuning~\citep{guo2026ctrl},
advantage estimation~\citep{yang2026rise}, and preference
optimization~\citep{hung2025nora}.
To improve the reliability of imagined feedback,
VLAW~\citep{guo2026vlaw} adapts its world model using real rollouts,
while WoVR~\citep{jiang2026wovr} regulates imagined rollouts and
co-evolves the world model and policy.
Our approach uses observed demonstration futures to repair
action-inconsistent predictions without additional environment
interaction. The repaired video expert then guides selective
negative fine-tuning of actions whose predicted progress falls
below the demonstrated reference.

%% file: sections/preliminary.tex
\section{Preliminaries}
\label{sec:prelim}

\textbf{World action model.}
To leverage world representations for action generation without explicit
future video generation at inference,
\method{} adopts the shared attention world action model architecture of
Fast-WAM~\citep{yuan2026fast}.  Each expert demonstration comprises
$x_i=(o_i,s_i,\ell_i)$, consisting of a current observation, robot state, and
language instruction, together with an expert action chunk $a_i^\star$ and its
observed future video $y_i^\star$.

The architecture comprises a video expert and an action expert, both
implemented as diffusion transformers.  The video stream contains the latent of the
current frame and noised future video latents, whereas the action stream
contains noised action chunks.  Within each MoT block, the two streams
interact through shared masked self-attention, followed by cross-attention
to state and language representations.  In our configuration, the video
expert also receives the action chunk.  The video expert defines the
action-conditioned video model
$W_\phi(y\mid x,a)$, and the action expert defines $\pi_\theta(a\mid x)$.  The
video expert therefore supplies world representations for action generation
and maps candidate actions to imagined consequences during post-training.

\paragraph{Flow matching initialization.}
We jointly train the video and action experts on demonstrations using
flow matching. Let $z_0$ denote an observed future video latent or a
demonstrated action chunk.
For $t\in[0,1]$ sampled from the training schedule and
$\epsilon\sim\mathcal N(0,I)$, we construct
$z_t=(1-t)z_0+t\epsilon$ with target velocity
$v^\star=\epsilon-z_0$.
The objective for expert $b\in\{\mathrm v,\mathrm a\}$ is
\begin{equation}
    \ell^b(z_0)
    =\mathbb E_{t,\epsilon}\!
    \left[
    \left\|v_{\psi_b}(z_t,t)-v^\star\right\|_2^2
    \right],
    \label{eq:fm}
\end{equation}
where $\psi_{\mathrm v}=\phi$ and $\psi_{\mathrm a}=\theta$.
Conditioning inputs are omitted for brevity. This objective
fits demonstration endpoints but does not compare them with endpoints sampled
from the current model.  A generated future may therefore fail to realize its
conditioning action, and a sampled action may produce an inferior imagined
consequence, without receiving a targeted corrective signal.

%% file: sections/method.tex
\section{Consequence-Grounded Post-Training}
\label{sec:method}

\begin{figure*}[t]
    \centering
    \includegraphics[width=\textwidth]{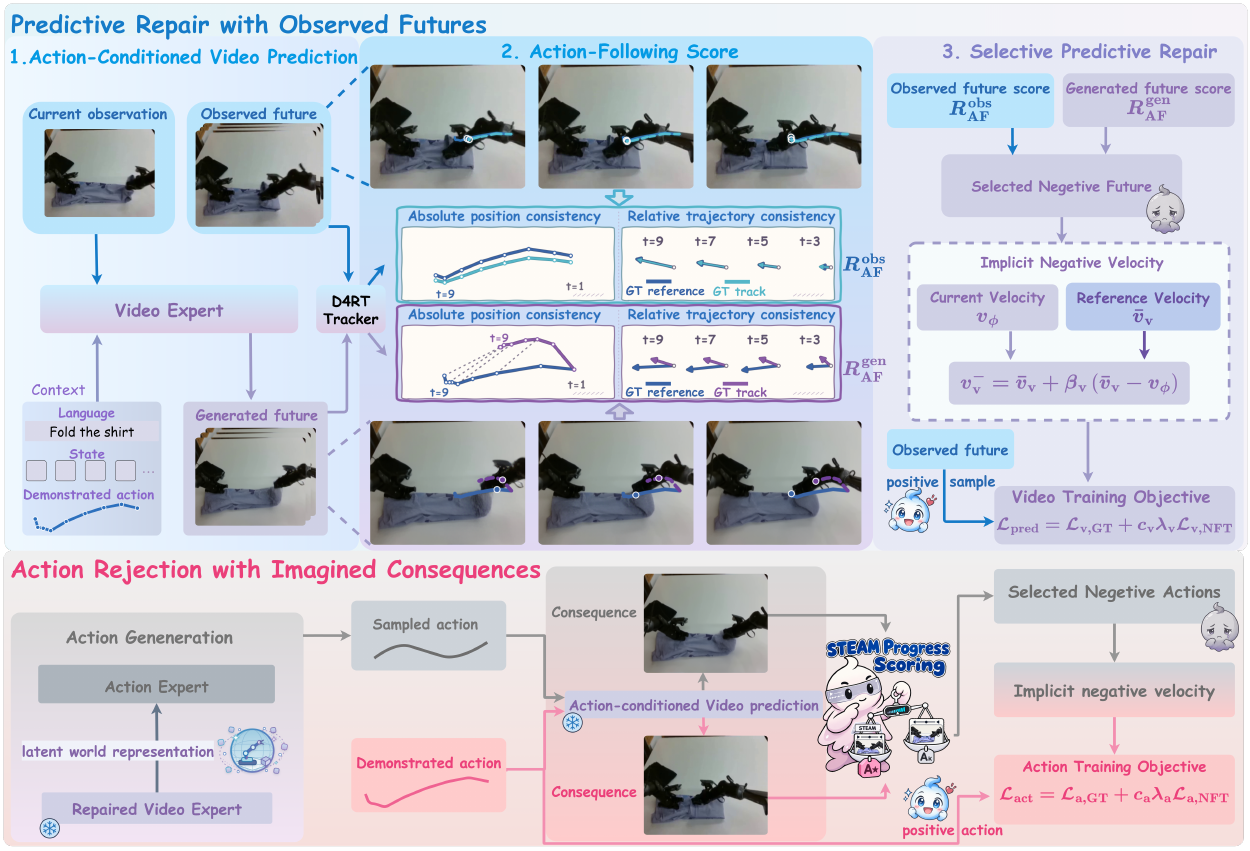}
    \caption{Overview of \method{}.
    Predictive repair compares generated robot motion with observed motion
    under the same demonstrated action. The repaired video expert is then
    frozen and predicts consequences for selecting inferior action samples.
    }
    \label{fig:pipeline}
\end{figure*}

\method{} refines a world action model through predictive repair
followed by action rejection (Figure~\ref{fig:pipeline}).
Predictive repair uses observed futures to identify and suppress
action-inconsistent video predictions
(Section~\ref{sec:predictive_repair}).
With the repaired video expert frozen, action rejection compares
the imagined progress of sampled and demonstrated actions to select
inferior action samples for negative fine-tuning
(Section~\ref{sec:action_repair}).

\subsection{Predictive Repair with Observed Futures}
\label{sec:predictive_repair}

Predictive repair uses the observed consequences of demonstrated actions
to identify predictions that fail to follow their conditioning actions.
For each demonstration $(x_i,a_i^\star,y_i^\star)$, the observed future
$y_i^\star$ provides a reference for $K_{\mathrm v}$ futures sampled under
the same action and initial observation:
\begin{equation}
    \hat y_{i,k}\sim W_\phi(\cdot\mid x_i,a_i^\star),
    \qquad k=1,\ldots,K_{\mathrm v}.
    \label{eq:video_samples}
\end{equation}
We assess action consistency in both observed and generated futures
through their agreement with the recorded robot motion.

\paragraph{Action-following score.}
Robot kinematics maps the demonstrated proprioceptive trajectory to
reference landmark positions $p_{i,j,f}$ in the camera frame, where $j$
indexes landmarks and $f$ indexes video frames. For a video $y$, a frozen
point tracker estimates landmark image locations and visibility
probabilities $q_{i,j,f}(y)\in[0,1]$. We obtain the tracked positions
$\hat p_{i,j,f}(y)$ in the same camera frame by lifting the pixels using
the kinematic reference depth and known camera intrinsics.
To restrict motion comparisons to reliable tracks, we evaluate visibility
over $\mathcal F_{i,j}$, the frames with an in-image reference projection
after the first such frame $f^{\mathrm{init}}_{i,j}$.
For nonempty $\mathcal F_{i,j}$, trajectory reliability is
\begin{equation}
    \rho_{i,j}(y)
    =\frac{1}{|\mathcal F_{i,j}|}
      \sum_{f\in\mathcal F_{i,j}}q_{i,j,f}(y).
    \label{eq:app_tracking_reliability}
\end{equation}
We introduce a video-level trackability indicator $C(y)$ to indicate
whether a video provides reliable landmark trajectories for motion
comparison. It equals $1$ when at least one landmark trajectory has a
nonempty comparison frame set $\mathcal F_{i,j}$ and mean visibility
$\rho_{i,j}(y)\ge0.5$, and $0$ otherwise.

We introduce two measures of action consistency: absolute-position
consistency evaluates agreement with the reference landmark positions,
while relative-trajectory consistency evaluates agreement with the
reference displacements.
Displacements from $f^{\mathrm{init}}_{i,j}$ are
\[
    \Delta p_{i,j,f}=p_{i,j,f}-p_{i,j,f^{\mathrm{init}}_{i,j}},
    \qquad
    \Delta\hat p_{i,j,f}(y)=
    \hat p_{i,j,f}(y)-\hat p_{i,j,f^{\mathrm{init}}_{i,j}}(y).
\]
The corresponding position and displacement errors are
\[
    e^{\mathrm{abs}}_{i,j,f}(y)=\|\hat p_{i,j,f}(y)-p_{i,j,f}\|_2,
    \qquad
    e^{\mathrm{rel}}_{i,j,f}(y)=
    \|\Delta\hat p_{i,j,f}(y)-\Delta p_{i,j,f}\|_2.
\]
For each component $g\in\{\mathrm{abs},\mathrm{rel}\}$, the trajectory
score $S^g_{i,j}(y)$ averages Gaussian alignment scores over frames using
visibility weights, with $\sigma_g$ setting the spatial tolerance.
For generated futures, $\bar S_i^g(y)$ denotes the reliability-weighted
mean of these scores over landmarks:
\begin{equation}
    S^g_{i,j}(y)
    =\frac{\sum_{f\in\mathcal F_{i,j}}q_{i,j,f}(y)
      \exp\!\left[-\frac{(e^g_{i,j,f}(y))^2}{2\sigma_g^2}\right]}
      {\sum_{f\in\mathcal F_{i,j}}q_{i,j,f}(y)},
    \quad
    \bar S_i^g(y)
    =\frac{\sum_{j:\rho_{i,j}(y)\ge0.5}\rho_{i,j}(y)S^g_{i,j}(y)}
    {\sum_{j:\rho_{i,j}(y)\ge0.5}\rho_{i,j}(y)}.
    \label{eq:app_action_following_components}
\end{equation}

For generated futures, the action-following score
$R_{\mathrm{AF}}^{\mathrm{gen}}(y,a_i^\star)$ averages the two aggregate
scores. The observed-reference score
$R_{\mathrm{AF}}^{\mathrm{obs}}(y_i^\star,a_i^\star)$ instead uses
selected reliable landmarks to reduce the influence of poor tracks.
Let $j_{\mathrm{abs}}^{\max}$ be the reliable landmark with the highest
$\rho_{i,j}(y_i^\star)$, and $j_{\mathrm{rel}}^{\max}$ the reliable
landmark with the highest $S^{\mathrm{rel}}_{i,j}(y_i^\star)$. Then
\begin{equation}
    R_{\mathrm{AF}}^{\mathrm{gen}}(y,a_i^\star)
    =\bigl(\bar S_i^{\mathrm{abs}}(y)+\bar S_i^{\mathrm{rel}}(y)\bigr)/2,
    \quad
    R_{\mathrm{AF}}^{\mathrm{obs}}(y_i^\star,a_i^\star)
    =\bigl(S^{\mathrm{abs}}_{i,j_{\mathrm{abs}}^{\max}}(y_i^\star)
    +S^{\mathrm{rel}}_{i,j_{\mathrm{rel}}^{\max}}(y_i^\star)\bigr)/2.
    \label{eq:app_generated_action_following_reward}
\end{equation}
Aggregating generated trajectories prevents one consistent landmark from
masking inconsistent motion in other reliable tracks.
These scores are evaluated only when $C(y)=1$.
Appendix~\ref{app:video_reward} specifies the landmarks, tracking model,
and coordinate lifting.

\paragraph{Observed-reference selection.}
When both futures are trackable, their action-following score gap is
\begin{equation}
    \Delta R_{i,k}
    =R_{\mathrm{AF}}^{\mathrm{obs}}(y_i^\star,a_i^\star)
     -R_{\mathrm{AF}}^{\mathrm{gen}}(\hat y_{i,k},a_i^\star).
    \label{eq:video_reward_gap}
\end{equation}
We assign each generated future a negative fine-tuning weight
$w^{\mathrm v}_{i,k}\in[0,1]$ based on the observed reference:
\begin{equation}
    w^{\mathrm v}_{i,k}=
    \begin{cases}
        0, & C(y_i^\star)=0,\\
        1, & C(y_i^\star)=1,\ C(\hat y_{i,k})=0,\\
        \operatorname{clip}\!\bigl(
        (\Delta R_{i,k}-m_{\mathrm v})/\tau_{\mathrm v},0,1
        \bigr),
        & C(y_i^\star)=C(\hat y_{i,k})=1.
    \end{cases}
    \label{eq:video_selection_weight}
\end{equation}
Here $m_{\mathrm v}$ is the minimum score gap for selecting a trackable
prediction, and $\tau_{\mathrm v}>0$ controls how its weight increases
beyond the margin. A reliable observed reference is required for negative
fine-tuning of generated futures.

\paragraph{Negative fine-tuning.}
Both stages use the same objective for expert $b\in\{\mathrm v,\mathrm a\}$.
Let $z_i^\star$ and $\hat z_{i,k}$ denote its demonstration and sampled
endpoints: video latents for $b=\mathrm v$ and action chunks for
$b=\mathrm a$.
Demonstration supervision is retained through the flow-matching objective
in Equation~\ref{eq:fm},
$\mathcal L_{b,\mathrm{GT}}=\frac{1}{N}\sum_{i=1}^N\ell^b(z_i^\star)$,
where $N$ is the number of demonstrations in the batch.
For generated samples selected as negatives, we adopt the implicit
negative parameterization of DiffusionNFT~\citep{zheng2026diffusionnft}:
\begin{equation}
    v_b^-(z_t,t)
    =(1+\beta_b)\bar v_b(z_t,t)-\beta_bv_{\psi_b}(z_t,t),
    \label{eq:implicit_negative}
\end{equation}
where $\bar v_b$ is a frozen copy of the corresponding velocity predictor
at the start of its post-training stage, $\beta_b>0$ controls the negative
parameterization, and $\psi_{\mathrm v}=\phi$, $\psi_{\mathrm a}=\theta$.
Negative fine-tuning regresses the implicit velocity $v_b^-$ toward
the flow targets of selected generated samples, weighted by $w^b_{i,k}$:
\begin{equation}
    \mathcal L_{b,\mathrm{NFT}}
    =\frac{\sum_{i,k}w^b_{i,k}
    \mathbb E_{t,\epsilon}\!\left[
    \left\|v_b^-(\hat z_{i,k,t},t)
    -v_{i,k}^\star\right\|_2^2\right]}
    {\sum_{i,k}w^b_{i,k}+\delta}.
    \label{eq:generic_negative}
\end{equation}
Here $\hat z_{i,k,t}=(1-t)\hat z_{i,k}+t\epsilon$ and
$v_{i,k}^\star=\epsilon-\hat z_{i,k}$ follow the flow-matching
construction in Equation~\ref{eq:fm}, and $\delta>0$ ensures a
well-defined loss when no negatives are selected.
The resulting post-training objective is
\begin{equation}
\mathcal L_b
=\mathcal L_{b,\mathrm{GT}}
+c_b\lambda_b\mathcal L_{b,\mathrm{NFT}}.
\label{eq:generic_total}
\end{equation}
Here $\lambda_b>0$ weights the negative loss, and $c_b\in[0,1]$
limits its contribution to a prescribed fraction of the
demonstration loss.
For predictive repair, the endpoints are observed and generated video
latents, with weights $w^{\mathrm v}_{i,k}$ from
Equation~\ref{eq:video_selection_weight}. We minimize
$\mathcal L_{\mathrm v}$ with respect to $\phi$, keeping the action expert
fixed. Appendix~\ref{app:video_nft} specifies the cap and optimization details.

\subsection{Action Rejection with Imagined Consequences}
\label{sec:action_repair}

\paragraph{Video representations and imagined consequences.}
Unlike demonstrated actions, newly sampled actions have no recorded
consequences. We therefore use the frozen, repaired video expert
$W_{\phi^\star}$ to evaluate their imagined outcomes while retaining
its role in providing world representations for action generation.
We sample $K_{\mathrm a}$ candidate actions from the current policy:
\begin{equation}
    \hat a_{i,k}\sim\pi_\theta(\cdot\mid x_i),
    \qquad k=1,\ldots,K_{\mathrm a}.
    \label{eq:action_samples}
\end{equation}
Let $\tilde y_i^\star$ and $\tilde y_{i,k}$ denote the futures predicted
by the repaired video expert under the demonstrated action $a_i^\star$
and sampled action $\hat a_{i,k}$, respectively, from the same initial
observation:
\begin{equation}
    \tilde y_i^\star\sim W_{\phi^\star}(\cdot\mid x_i,a_i^\star),
    \qquad
    \tilde y_{i,k}\sim W_{\phi^\star}(\cdot\mid x_i,\hat a_{i,k}).
    \label{eq:action_rollouts}
\end{equation}

\paragraph{Consequence-based action selection.}
We evaluate task progress in predicted futures using a frozen
temporal progress model following STEAM~\citep{liu2026steam}.
Its score $A_{\mathrm{STEAM}}(x_i,y)$ measures progress from the current
observation to the final frame of $y$, conditioned on the language
instruction.
We define the progress gap $\Delta A_{i,k}$ as the difference between
the predicted progress of the demonstrated and sampled actions:
\begin{equation}
    \Delta A_{i,k}
    =A_{\mathrm{STEAM}}(x_i,\tilde y_i^\star)
     -A_{\mathrm{STEAM}}(x_i,\tilde y_{i,k}).
    \label{eq:relative_gap}
\end{equation}
A positive $\Delta A_{i,k}$ indicates lower predicted progress for the
sampled action.
Actions whose predicted progress falls sufficiently below the
demonstrated reference receive negative fine-tuning weights
$w^{\mathrm a}_{i,k}\in[0,1]$:
\begin{equation}
    w^{\mathrm a}_{i,k}
    =\operatorname{clip}\!\bigl(
    (\Delta A_{i,k}-m_{\mathrm a})/\tau_{\mathrm a},0,1
    \bigr).
    \label{eq:action_reward_gate}
\end{equation}
Here $m_{\mathrm a}$ is the rejection margin, and
$\tau_{\mathrm a}>0$ controls the increase in weight beyond it.
Appendix~\ref{app:steam_model} details the scorer training.

\paragraph{Rejection-only action fine-tuning.}
We apply Equation~\ref{eq:generic_total} with $b=\mathrm a$, using
demonstrated actions $a_i^\star$ as positive endpoints and sampled actions
$\hat a_{i,k}$ as negative candidates weighted by $w^{\mathrm a}_{i,k}$
from Equation~\ref{eq:action_reward_gate}.
The frozen reference $\bar v_{\mathrm a}$ is initialized from the action
expert at the start of this stage. We optimize $\theta$ while keeping
the repaired video expert and STEAM scorer fixed.

%% file: sections/experiments.tex
\section{Experiments}
\label{sec:experiments}

Our experiments address five questions.
\textbf{(Q1) Simulation performance:} how does \method{} compare with
recent vision language action policies and world action models on
RoboTwin~2.0~\citep{chen2025robotwin}?
\textbf{(Q2) Negative selection:} does predictive repair improve the
agreement between negative selections based on imagined consequences
and STEAM judgments of actual executions?
\textbf{(Q3) Ablation studies:} which post-training components and
action training objectives account for the gains in task success?
\textbf{(Q4) Real-world effectiveness:} does \method{} improve task
completion in long-horizon real-world manipulation?
\textbf{(Q5) Behavioral analysis:} do imagined consequences reveal the
lack of progress observed during shirt folding?

\subsection{Simulation on RoboTwin 2.0}
\label{sec:sim_robotwin}

\paragraph{Experimental protocol.}
We follow the RoboTwin~2.0 multitask protocol used by Fast-WAM.  The training
set combines 2,500 demonstrations collected in clean scenes and 25,000
demonstrations collected with scene randomization, spanning 50 bimanual
manipulation tasks.  We evaluate every task in both clean and randomized
environments using 100 closed-loop trials per setting, and report success
rates averaged over the 50 tasks.

Our base model is Fast-WAM~\citep{yuan2026fast} with an
action-conditioned video expert.
We first train both experts jointly with flow matching for 5 epochs.
Action-conditioned Fast-WAM continues joint flow matching for
2 additional epochs, whereas \method{} applies predictive repair followed by action
post-training, training each expert for 2 epochs.
Appendix~\ref{app:simulation_training} details the ablation protocol, and
Appendix~\ref{app:implementation} reports the optimization and sampling settings.

\begin{table*}[t]
    \centering
    \caption{Closed-loop success rates (\%) on RoboTwin~2.0.}
    \label{tab:robotwin_main}
    \small
    \setlength{\tabcolsep}{11pt}
    \begin{tabular}{lccc}
        \toprule
        Method & Clean & Randomized & Average \\
        \midrule
        \multicolumn{4}{l}{Vision language action policies} \\
        $\pi_0$~\citep{black2024pi_0}           & 65.9 & 58.4 & 62.2 \\
        X-VLA~\citep{zheng2026x}                & 72.9 & 72.8 & 72.9 \\
        $\pi_{0.5}$~\citep{intelligence2025pi_} & 82.7 & 76.8 & 79.8 \\
        ABot-M0~\citep{yang2026abot}            & 81.2 & 80.4 & 80.8 \\
        LingBot-VLA~\citep{wu2026pragmatic}     & 86.5 & 85.3 & 85.9 \\
        HoloBrain-0-QW~\citep{lin2026holobrain}  & 91.9 & 92.3 & 92.1 \\
        \midrule
        \multicolumn{4}{l}{World action models} \\
        Motus~\citep{Bi_2026_CVPR}              & 88.7 & 87.0 & 87.9 \\
        GigaWorld-Policy~\citep{ye2026gigaworld}  & 85.6 & 85.3 & 85.5 \\
        LingBot-VA~\citep{li2026causal}          & 92.9 & 91.6 & 92.3 \\
        Fast-WAM~\citep{yuan2026fast}            & 91.9 & 91.8 & 91.9 \\
        
        \midrule
        \multicolumn{4}{l}{Our evaluations} \\
        Action-conditioned Fast-WAM       & 90.6 & 90.4 & 90.5 \\
        \method                              & \textbf{93.8} & \textbf{93.7} & \textbf{93.8} \\
        \bottomrule
    \end{tabular}
\end{table*}

\paragraph{Main comparison.}
To answer Q1, we compare \method{} with recent vision-language
action policies and world action models
(Table~\ref{tab:robotwin_main}).
\method{} achieves the highest success rates in both evaluation settings,
reaching 93.8\% in clean environments and 93.7\% under scene randomization.
The gains over LingBot-VA in clean environments and HoloBrain-0-QW
under randomization are 0.9 and 1.4 percentage points, respectively.
Relative to action-conditioned Fast-WAM,
\method{} improves success by 3.2 and 3.3 percentage points in clean
and randomized environments, respectively.

\subsection{Action Negative Selection}
\label{sec:negative_selection}

To answer Q2, we test whether predictive repair improves the selection of action negatives
from imagined consequences on evaluation episodes.
Across 500 contexts from all 50 RoboTwin tasks,
we predict consequences for 2,000 shared candidate actions using video experts
trained with continued flow matching or predictive repair.
Each candidate and its expert reference
are also executed from the same simulator state. The same frozen STEAM
evaluator scores actual and imagined consequences, providing reference
labels from actual executions and predicted selections from imagined futures.
These simulator executions are used only for evaluation, not for post-training.
We evaluate margins of 0.02, 0.05, and 0.08, applying the same margin to
actual and imagined consequence gaps in each comparison. Precision and
recall are computed from pooled counts across the 2,000 candidates.
Predictive repair improves negative-selection precision and recall
at all three margins (Table~\ref{tab:negative_selection}).
At the training margin of 0.08, precision increases from 64.6\%
to 69.8\%, while recall rises from 35.5\% to 50.6\%.
With candidate actions and the STEAM evaluator held fixed, the
improved agreement with execution-based judgments suggests that
predictive repair provides more reliable predictions for selecting
action negatives.
Appendix~\ref{app:negative_selection} provides the protocol and complete counts.

\begin{table}[t]
    \centering
    \caption{Action negative-selection precision and recall (\%) at three
    margins. Reference labels are obtained by applying each margin to
    STEAM score gaps from actual executions; predicted selections use
    the same margin on imagined consequences.}
    \label{tab:negative_selection}
    \small
    \setlength{\tabcolsep}{7pt}
    \begin{tabular}{@{}lcccc@{}}
        \toprule
        & \multicolumn{2}{c}{Precision} & \multicolumn{2}{c}{Recall} \\
        \cmidrule(lr){2-3}\cmidrule(lr){4-5}
        Margin & Continued FM & Repair & Continued FM & Repair \\
        \midrule
        0.02 & 76.9 & \textbf{79.6} & 44.4 & \textbf{63.7} \\
        0.05 & 63.2 & \textbf{76.8} & 38.7 & \textbf{56.7} \\
        0.08 & 64.6 & \textbf{69.8} & 35.5 & \textbf{50.6} \\
        \bottomrule
    \end{tabular}
\end{table}

\subsection{Ablation Studies}
\label{sec:exp_ablation}
\label{sec:exp_coupling}

\paragraph{Contributions of the two stages.}
To answer Q3, we first examine the contributions of predictive
repair and action rejection
(Table~\ref{tab:predictive_action_coupling}). All four variants train the video expert first, followed by
the action expert. GT FM denotes flow matching on
demonstrations alone. Repair adds negative fine-tuning of generated videos
selected against observed futures, whereas Rejection adds negative
fine-tuning of sampled actions selected by their imagined progress.

\FloatBarrier
\begin{table}[!htbp]
    \centering
    \begin{minipage}[t]{0.49\textwidth}
    \vspace{0pt}
    \centering
    \caption{Component ablation. Success rates (\%).}
    \label{tab:predictive_action_coupling}
    \fontsize{8}{10}\selectfont
    \setlength{\tabcolsep}{2.5pt}
    \begin{tabular}{@{}llcc@{}}
        \toprule
        Video & Action & Clean & Rand. \\
        \midrule
        GT FM  & GT FM     & 90.8 & 90.5  \\
        GT FM  & Rejection & 91.1 & 90.7 \\
        Repair & GT FM     & 91.9 & 91.4 \\
        Repair & Rejection & \textbf{93.8} & \textbf{93.7} \\
        \bottomrule
    \end{tabular}
    \end{minipage}\hfill
    \begin{minipage}[t]{0.48\textwidth}
    \vspace{0pt}
    \centering
    \caption{Action objectives. Success rates (\%).}
    \label{tab:action_objective_ablation}
    \fontsize{8}{10}\selectfont
    \setlength{\tabcolsep}{2.5pt}
    \begin{tabular}{@{}lcc@{}}
        \toprule
        Variant & Clean & Rand. \\
        \midrule
        All negatives                           & 86.8 & 85.9 \\
        Random negatives                        & 87.1 & 87.2 \\
        RECAP + STEAM                           & 92.6 & 92.3 \\
        \method{} (ours)                         & \textbf{93.8} & \textbf{93.7} \\
        \bottomrule
    \end{tabular}
    \end{minipage}
\end{table}

With the action objective fixed to GT FM, predictive repair increases
success from 90.8\% to 91.9\% in clean environments and from 90.5\%
to 91.4\% under randomization, supporting its role in learning more
useful world representations for action generation.
Beyond representation learning, repair also strengthens the gains
from action rejection: improvements of 0.3 and 0.2 percentage points
without repair increase to 1.9 and 2.3 points after repair in clean
and randomized environments, respectively.
Together with the improved negative-selection accuracy in
Table~\ref{tab:negative_selection}, these results support predictive
repair as a source of more reliable consequence feedback for
negative action fine-tuning.

\paragraph{Action training objective.}
We next compare action training objectives with the repaired
video expert fixed (Table~\ref{tab:action_objective_ablation}).
All negatives treats every policy sample as a negative, while Random negatives
selects each sampled action independently with probability 0.5.
Both retain demonstration flow matching.
RECAP + STEAM~\citep{intelligence2025pi} conditions the policy on a
binary advantage indicator derived from STEAM estimates.

Rejecting all samples or a random subset underperforms demonstration
flow matching alone in both settings
(Repair / GT FM in Table~\ref{tab:predictive_action_coupling}),
showing that poorly selected negatives can degrade the policy despite
retaining demonstration supervision. Consequence-based rejection instead
outperforms RECAP + STEAM by 1.2 and 1.4 percentage points in clean and
randomized environments, respectively. These results support selective
rejection as a more effective use of predicted progress than advantage
conditioning in this setting.

\FloatBarrier
\subsection{Real-World Evaluation}
\label{sec:real_robot}

\paragraph{Robot setup and tasks.}
We evaluate on a PIPER platform with two arms, equipped with one front camera
and two RGB cameras mounted on the wrists.  The policy receives the three camera
views and the current robot state, and predicts bimanual end-effector action
chunks.  We evaluate two long-horizon tasks covering deformable manipulation
and tabletop cleanup.  In Fold Shirt, the robot transforms a shirt into a
rectangular configuration through an extended sequence of coordinated bimanual
actions.  In Clean Table, it repeatedly selects, grasps, and places
bottles and bottle caps from the tabletop into a trash bin.  Hardware, data, and deployment details appear in
Appendix~\ref{app:real_robot_setup}.

\paragraph{Baselines.}
We compare against 4 baselines: $\pi_{0.5}$~\citep{intelligence2025pi_},
X-VLA~\citep{zheng2026x}, Motus~\citep{Bi_2026_CVPR}, and
Fast-WAM~\citep{yuan2026fast}.  Each baseline is trained for 7 epochs on the same demonstrations
for each task, using its original architecture, observation processing, and
action representation.  We initialize \method{} from Fast-WAM after 5 epochs
and train each expert for 2 additional epochs using its corresponding
post-training objective.

\begin{figure*}[t]
    \centering
    \includegraphics[width=\textwidth]{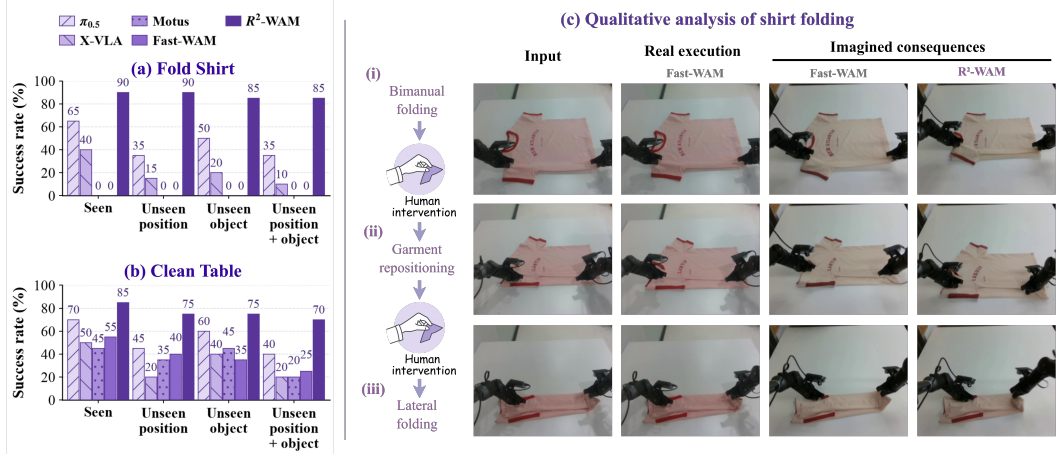}
\caption{Real-world evaluation and qualitative analysis.
(a--b) Success rates on Fold Shirt and Clean Table.
(c) Fast-WAM executions and imagined consequences of actions from
Fast-WAM and \method{} during (i) bimanual folding,
(ii) garment repositioning, and (iii) lateral folding.
Human intervention in (c) enables later-stage analysis;
these rollouts are excluded from success rates.}
    \label{fig:real_robot_main}
\end{figure*}

\paragraph{Evaluation protocol.}
We organize evaluation as a 2$\times$2 design over initial position and object
instance.  Seen uses the training position region and familiar objects.
Unseen Position and Unseen Object vary one factor at a time, while
Unseen Position + Object varies both.  Each condition contains 20
autonomous trials per method and task.  Each trained policy is evaluated across
all 4 conditions.  Figure~\ref{fig:real_robot_main}(a--b)
reports success for each task; complete numerical results, Wilson
confidence intervals, and the complete protocol are provided in
Appendices~\ref{app:real_robot_protocol}
and~\ref{app:real_robot_statistics}.

\paragraph{Real-world performance.}
To answer Q4, we compare task success on Fold Shirt and
Clean Table (Figure~\ref{fig:real_robot_main}(a--b)).
On Fold Shirt, a long-horizon bimanual task, \method{} achieves
87.5\% average success, whereas neither Fast-WAM nor Motus completes
the task in any evaluation trial.
The VLA baselines, $\pi_{0.5}$ and X-VLA, achieve higher success
than these WAM baselines but remain below \method{} across all
four evaluation conditions.
On Clean Table, \method{} also outperforms all evaluated baselines
across the four conditions. Under joint variation, it achieves 70.0\%
success, exceeding the strongest baseline, $\pi_{0.5}$, by
30 percentage~points.

\paragraph{Qualitative analysis of shirt folding.}
To answer Q5, we compare Fast-WAM executions with imagined
consequences at three stages of shirt folding
(Figure~\ref{fig:real_robot_main}(c)).
At each stage, Fast-WAM establishes a grasp or contact but
does not complete the subsequent garment motion. During bimanual folding, it lifts the garment
without folding it forward.  During repositioning, it contacts the garment
without pulling it toward the robot.  During lateral folding, it holds and
lifts a corner without completing the fold.

For each initial observation, we use the same repaired video expert to
predict the consequences of actions sampled from Fast-WAM and \method{}.
The predictions for Fast-WAM actions show a similar lack of progress to the
displayed executions, whereas those for \method{} depict forward folding,
garment pulling, and lateral folding, respectively.
These examples illustrate the stalled behaviors that consequence
feedback is intended to correct: grasping or lifting the garment
without advancing the folding sequence.
Their appearance in the predicted futures suggests that the repaired
video expert can help identify such actions for rejection during
post-training, with the aim of improving completion of the full task.

%% file: sections/conclusion.tex
\section{Conclusion}
\label{sec:conclusion}

We presented \method{}, a repair-and-reject framework for
post-training world action models.
Predictive repair uses observed demonstration futures to improve
the action fidelity of video predictions.
The repaired video expert then guides negative fine-tuning of
sampled actions whose predicted progress falls below the
demonstrated reference.
Experiments on RoboTwin~2.0 and two real-world manipulation tasks
show improved task completion, including long-horizon shirt
folding that Fast-WAM fails to complete.
Further analysis shows that predictive repair improves
negative-selection accuracy, while action rejection yields
larger gains in task success with the repaired video expert.

%% file: sections/method_appendix.tex
\section{Details of Predictive Repair}
\label{app:predictive_grounding}

\subsection{Action-Following Score}
\label{app:video_reward}

Section~\ref{sec:predictive_repair} defines the geometric scores, tracking
reliability, and aggregation rules. Here we specify how the tracked image
locations are expressed in the reference camera coordinates.
Let $\hat{\mathbf u}_{i,j,f}(y)$ be the tracked pixel location of landmark
$j$ in frame $f$, $d_{i,j,f}$ its depth from the demonstrated robot
kinematics, and $\mK$ the fixed camera intrinsic matrix. We use
\begin{equation}
    \hat p_{i,j,f}(y)
    =d_{i,j,f}\mK^{-1}
      \begin{bmatrix}\hat{\mathbf u}_{i,j,f}(y)\\1\end{bmatrix}.
    \label{eq:app_landmark_lifting}
\end{equation}
Observed and generated videos use the same reference depths and camera
coordinates. The Gaussian bandwidths in
Equation~\ref{eq:app_action_following_components} are
$\sigma_{\mathrm{abs}}=\sigma_{\mathrm{rel}}=0.05\,\mathrm m$.

\paragraph{Tracking model and landmarks.}
We use OpenD4RT,\footnote{\url{https://github.com/Lijiaxin0111/Open-d4rt}}
an unofficial implementation of D4RT~\citep{zhang2026efficiently},
as a frozen point tracker. It processes the head-camera view of each
9-frame clip, comprising the current frame and 8 future frames, at
256$\times$256 resolution; the video expert receives all three views.

For RoboTwin, we place three virtual landmarks in the local coordinate
frame of link 6 on each arm:
\[
    (0,\,0,\,0), \qquad
    (0.065,\,0.052,\,0), \qquad
    (0.065,\,-0.052,\,0).
\]
All coordinates are in meters, giving six landmarks across the two arms.
OpenD4RT returns image trajectories and visibility logits; their sigmoid
probabilities provide $q_{i,j,f}$ in
Equations~\ref{eq:app_tracking_reliability}
and~\ref{eq:app_action_following_components}.

\subsection{Observed-Reference Negative Selection}
\label{app:video_selection}

Equation~\ref{eq:video_selection_weight} specifies the selection and weighting
rule, including the cases where the observed or generated future cannot be
reliably tracked. We use $m_{\mathrm v}=0.02$ and
$\tau_{\mathrm v}=0.01$ in all experiments.
Thus, for two trackable futures, a generated sample contributes to the
negative loss only when $\Delta R_{i,k}>0.02$, and its weight reaches
one when $\Delta R_{i,k}\ge0.03$.
The observed future contributes to the demonstration loss regardless
of its tracking reliability.

\subsection{Forward-Process Negative Fine-Tuning}
\label{app:video_nft}

The predictive objective combines the demonstration loss defined in
Section~\ref{sec:predictive_repair} with the weighted negative loss in
Equation~\ref{eq:generic_negative} with $b=\mathrm v$.
For each generated latent endpoint,
the noised latent and flow target are
\[
    \hat z_{i,k,t}=(1-t)\hat z_{i,k}+t\epsilon,
    \qquad v_{i,k}^\star=\epsilon-\hat z_{i,k},
    \qquad \epsilon\sim\mathcal N(0,I).
\]
The negative loss fits the implicit velocity $v_{\mathrm v}^-$ rather than
the trainable velocity directly. The coefficient of the latter in
Equation~\ref{eq:implicit_negative} is $-\beta_{\mathrm v}$.
Sampled endpoints and selection weights are held fixed during each update.

For both experts $b\in\{\mathrm v,\mathrm a\}$, the coefficient in
Equation~\ref{eq:generic_total} is
\begin{equation}
    c_b=\min\!\left(
    1,\frac{r_{\max}^b\mathcal L_{b,\mathrm{GT}}}
    {\lambda_b\mathcal L_{b,\mathrm{NFT}}+\delta}\right),
    \label{eq:app_loss_cap}
\end{equation}
where $r_{\max}^b$ is the maximum ratio of the weighted negative loss
contribution to the demonstration loss. We hold $c_b$ constant during
differentiation. The constant $\delta>0$ stabilizes both the cap and the
negative-loss normalization; the negative loss is zero when all sample
weights are zero.
We use $\beta_{\mathrm v}=0.1$, $\lambda_{\mathrm v}=0.1$,
$r_{\max}^{\mathrm v}=0.25$, and $\delta=10^{-8}$.
Both stages use the shifted flow-time schedule inherited from Fast-WAM:
we draw $u\sim\mathcal U(0,1)$ and set $t=5u/(1+4u)$.
The frozen reference is initialized once at the start of predictive repair.
Appendix~\ref{app:frozen_reference_characterization} analyzes the
negative objective under a fixed reference.

\section{Details of Action Rejection}
\label{app:action_rejection}

\subsection{Forward-Process Negative Fine-Tuning}
\label{app:action_nft}

Action post-training uses Equation~\ref{eq:generic_total} with
$b=\mathrm a$, demonstration endpoints $z_i^\star=a_i^\star$, and
sampled endpoints $\hat z_{i,k}=\hat a_{i,k}$.
The negative loss follows Equation~\ref{eq:generic_negative}, with
weights from Equation~\ref{eq:action_reward_gate}.
We use $m_{\mathrm a}=0.08$ and $\tau_{\mathrm a}=0.10$ in simulation,
and $m_{\mathrm a}=0.05$ and $\tau_{\mathrm a}=0.04$ on the real robot.
The remaining coefficients are $\beta_{\mathrm a}=0.1$,
$\lambda_{\mathrm a}=0.2$, and $r_{\max}^{\mathrm a}=0.15$ in both domains.
The numerical stability constant, flow-time schedule, and treatment of
sampled endpoints and weights are the same as in predictive repair.
The loss cap follows Equation~\ref{eq:app_loss_cap}.
The action reference is initialized once from the
action expert at the start of this stage and remains frozen.

\subsection{Temporal Progress Model}
\label{app:steam_model}

Following STEAM~\citep{liu2026steam}, we learn task-conditioned temporal
progress from pairs of multi-view observations and their language
instruction. SigLIP2 so400m-patch14-384 and Gemma-3 270M encode the visual
and language inputs, respectively. A 32-way classifier predicts signed
temporal displacement from forward and reversed frame pairs, with the
expected displacement normalized to $[-1,1]$. Both encoders are trained
jointly with the projection layers, multimodal fusion layers, and
temporal classification head.

In simulation, training uses demonstrations from all 50 RoboTwin tasks,
comprising 2,500 clean and 25,000 randomized episodes. For real-world
evaluation, we train separate scorers for Fold Shirt and Clean Table
using the same task demonstrations as policy training
(Table~\ref{tab:real_robot_data}). Frame pairs are sampled with a maximum
temporal separation of $k$ frames: $k=32$ for RoboTwin and Fold Shirt,
and $k=64$ for Clean Table. The classifier retains 32 temporal bins in
all settings.

All scorers share the same optimization settings. Images are resized to
384$\times$384, and each predictor is trained for 16,000 steps using
AdamW with learning rate $5\times10^{-5}$, weight decay $10^{-5}$,
betas (0.9, 0.95), and a global batch size of 512 on 8 NVIDIA A800 GPUs.
The learning rate is shared by the encoders and prediction head.
Following STEAM, we train an ensemble of three predictors and take
the minimum of their progress estimates as $A_{\mathrm{STEAM}}$.
Positive scores indicate progress and negative scores indicate regression.
The ensemble remains frozen during WAM post-training and evaluates
progress between the current observation and the final frame of each
imagined future.

\section{Implementation Details}
\label{app:implementation}

\subsection{Optimization and Sampling}
\label{app:optimization}

Post-training starts from a Fast-WAM model trained for 5 epochs by
joint video and action flow matching. For real-world experiments, we
follow the Fast-WAM initialization: the video DiT, text encoder, and
video VAE are initialized from pretrained Wan2.2-5B, and the action
expert is initialized according to the Fast-WAM configuration.
The initial 5 epochs use real-world demonstrations, without transferring
a policy checkpoint trained on RoboTwin. The video expert is then refined
for 2 epochs.  We then freeze the repaired video expert and train the
action expert for 2 epochs.  Both stages use AdamW with learning rate
$10^{-4}$, betas (0.9, 0.95),
and zero weight decay.  The learning rate follows a linear warmup over the
first 5\% of training and cosine decay to $10^{-6}$.  Each stage is trained on
16 NVIDIA A800 GPUs.  The predictive and action objectives update approximately
5.00B and 1.02B parameters, respectively.
These optimization settings apply to both simulation and real-world
post-training.

Predictive repair samples $K_{\mathrm v}$ futures for each demonstration
context.  Action post-training samples $K_{\mathrm a}$ action chunks and rolls
out each candidate with the frozen video expert.  We use 5-step Euler
integration with the shifted flow schedule for both video and action sampling.
At inference, following Fast-WAM, the action
expert uses 10 Euler steps to generate a 32-step action chunk and executes its
first 24 steps before generating the next chunk.

\subsection{Post-Training Hyperparameters}
\label{app:post_training_hyperparameters}

Table~\ref{tab:post_training_hyperparameters} summarizes the coefficients used
in simulation and on the real robot.  Predictive repair uses the same settings
in both domains.  For action post-training, the margin and weighting scale
differ between domains; all remaining coefficients are shared.

\begin{table}[h]
    \centering
    \caption{Post-training hyperparameters in simulation and on the real robot.}
    \label{tab:post_training_hyperparameters}
    \small
    \setlength{\tabcolsep}{5pt}
    \begin{tabular}{lcc}
        \toprule
        Parameter & Simulation & Real robot \\
        \midrule
        $K_{\mathrm v}$ & 4 & 4 \\
        $m_{\mathrm v}$ & 0.02 & 0.02 \\
        $\tau_{\mathrm v}$ & 0.01 & 0.01 \\
        $\beta_{\mathrm v}$ & 0.1 & 0.1 \\
        $\lambda_{\mathrm v}$ & 0.1 & 0.1 \\
        $r_{\max}^{\mathrm v}$ & 0.25 & 0.25 \\
        \midrule
        $K_{\mathrm a}$ & 4 & 4 \\
        $m_{\mathrm a}$ & 0.08 & 0.05 \\
        $\tau_{\mathrm a}$ & 0.10 & 0.04 \\
        $\beta_{\mathrm a}$ & 0.1 & 0.1 \\
        $\lambda_{\mathrm a}$ & 0.20 & 0.20 \\
        $r_{\max}^{\mathrm a}$ & 0.15 & 0.15 \\
        \bottomrule
    \end{tabular}
\end{table}

\section{Analysis of the Rejection Objective under a Fixed Reference}
\label{app:frozen_reference_characterization}

We provide a pointwise population analysis of the uncapped objective under a
fixed reference predictor and a fixed distribution of selected negatives.  The
analysis concerns the predictor output at a single input; it does not establish
convexity in neural network parameters or characterize training dynamics under
repeated negative sampling.

Fix a branch $b\in\{\mathrm v,\mathrm a\}$ and a complete predictor input
$\xi$, including the noised endpoint, flow time, and the conditioning of that
branch.  We write
\[
    v=v_{\psi_b}(\xi),\qquad
    \bar v=\bar v_b(\xi),\qquad
    \lambda=\lambda_b,\qquad
    \beta=\beta_b.
\]
Let $p_{\mathrm{GT}}(\xi)$ denote the density of noised inputs induced by
demonstration endpoints, and let $p_{\mathrm{neg}}(\xi)$ denote the weighted
density induced by selected negative endpoints.  The latter incorporates the
nonnegative selection weights used by the corresponding branch.  Their
conditional mean flow targets are
\begin{equation}
    m_{\mathrm{GT}}(\xi)
    =\mathbb E[v^\star\mid \xi,\mathrm{GT}],
    \qquad
    m_{\mathrm{neg}}(\xi)
    =\mathbb E[v^\star\mid \xi,\mathrm{neg}].
\end{equation}

Suppressing the dependence on $\xi$, the bias--variance decomposition of the
squared flow-matching loss gives the following objective, up to terms
independent of $v$:
\begin{equation}
    F_\xi(v)
    =
    p_{\mathrm{GT}}\|v-m_{\mathrm{GT}}\|_2^2
    +\lambda p_{\mathrm{neg}}
    \left\|(1+\beta)\bar v-\beta v-m_{\mathrm{neg}}\right\|_2^2.
    \label{eq:app_pointwise_objective}
\end{equation}
Define the effective regression target
\begin{equation}
    \widetilde{m}_{\mathrm{neg}}
    =\bar v+\frac{1}{\beta}
    \left(\bar v-m_{\mathrm{neg}}\right).
    \label{eq:app_effective_negative_target}
\end{equation}
Since
\[
    (1+\beta)\bar v-\beta v-m_{\mathrm{neg}}
    =-\beta(v-\widetilde{m}_{\mathrm{neg}}),
\]
Equation~\ref{eq:app_pointwise_objective} is equivalent to
\begin{equation}
    F_\xi(v)
    =p_{\mathrm{GT}}\|v-m_{\mathrm{GT}}\|_2^2
     +\lambda\beta^2p_{\mathrm{neg}}
      \|v-\widetilde{m}_{\mathrm{neg}}\|_2^2.
\end{equation}
Whenever
$p_{\mathrm{GT}}+\lambda\beta^2p_{\mathrm{neg}}>0$, the objective is strictly
convex in the pointwise function value, with
\begin{equation}
    \nabla_v^2F_\xi
    =2\left(p_{\mathrm{GT}}+\lambda\beta^2p_{\mathrm{neg}}\right)I
    \succ 0.
\end{equation}
Its unique minimizer is
\begin{equation}
    v_{\mathrm{opt}}(\xi)
    =
    \frac{
    p_{\mathrm{GT}}m_{\mathrm{GT}}
    +\lambda\beta^2p_{\mathrm{neg}}\widetilde{m}_{\mathrm{neg}}
    }{
    p_{\mathrm{GT}}+\lambda\beta^2p_{\mathrm{neg}}
    }.
    \label{eq:app_pointwise_solution}
\end{equation}
Because
\[
    \widetilde{m}_{\mathrm{neg}}-\bar v
    =\frac{\bar v-m_{\mathrm{neg}}}{\beta},
\]
$\widetilde{m}_{\mathrm{neg}}$ extrapolates from the fixed reference in the direction
opposite to the conditional mean target of the selected negatives.  The
pointwise optimum therefore balances the demonstration flow target with this
effective regression target.  When $p_{\mathrm{neg}}=0$ and
$p_{\mathrm{GT}}>0$, it reduces to the demonstration optimum
$m_{\mathrm{GT}}$.

This analysis differs from the reinforcement guidance derivation of
DiffusionNFT~\citep{zheng2026diffusionnft}.  DiffusionNFT obtains its positive
and negative distributions by reward reweighting samples from a single
reference policy, such that their mixture recovers that policy.  In our
setting, positive endpoints come from demonstrations, whereas negative
endpoints are selected from samples generated by the current model; the two distributions do
not form an analogous decomposition.  Equation~\ref{eq:app_pointwise_solution}
therefore characterizes the minimizer of our rejection objective rather than a
direction of reward improvement.  It neither guarantees higher environment
reward nor implies closed-loop policy improvement.

%% file: sections/experiments_appendix.tex
\FloatBarrier
\section{Simulation Experiment Details}
\label{app:simulation}

\subsection{Ablation Protocol}
\label{app:simulation_training}

The Fast-WAM implementation supports an action-conditioned video expert,
whereas its published results use action-unconditioned video prediction.
All ablations start from the same action-conditioned Fast-WAM checkpoint
trained for 5 epochs.  Optimization and sampling settings follow
Appendix~\ref{app:implementation}.

\paragraph{Predictive and action objectives.}
In
Table~\ref{tab:predictive_action_coupling}, each predictive objective is
applied for 2 epochs, followed by 2 epochs with the corresponding action
objective.  When predictive repair or action rejection is omitted, the
respective expert instead receives 2 additional epochs of flow matching on the
demonstrations.  Every setting therefore uses the same data and number of
optimizer updates.

\paragraph{Action objective ablation.}
Table~\ref{tab:action_objective_ablation} isolates the effect of the action
objective using a common frozen video expert after repair.  All variants
share the action initialization, data order, and training duration.  Variants
that use sampled actions also share the candidate sampling budget.
For all-sample rejection, every sampled action receives weight 1.  For random
selection, each sampled action independently receives weight 1 with probability
0.5 and weight 0 otherwise.  Both variants use the normalization in
Equation~\ref{eq:generic_negative} with $b=\mathrm a$, so the negative loss is averaged over
the selected samples in the minibatch.  It is zero when no sample is selected.

\paragraph{Action and video representations.}
In RoboTwin, each action is a 32-step sequence of 14 absolute joint and
gripper targets, comprising six arm joints and one gripper command per arm.
The demonstrations are recorded at 50~Hz.  Each video clip contains the current
frame and 8 predicted future frames, with the three camera views arranged
at a combined resolution of 384$\times$320.  The current frame provides visual
conditioning and is excluded from the flow-matching loss.  Consecutive frames
are separated by 4 action steps, so the clip spans the complete action chunk.

\subsection{Evaluation of Action-Conditioned Prediction}
\label{app:predictive_diagnostics}

\paragraph{Experimental setup.}
We compare five objectives for post-training the video expert.
Continued flow matching serves as a control with the same training data and
number of optimizer updates, but uses only the observed futures.
Repair with PSNR compares generated and observed head-camera images,
averaging PSNR over the 8 future frames.  Generated futures with an
average PSNR below 26~dB are selected as negatives and assigned
weight one in the negative fine-tuning objective.
We then evaluate the complete action-following score and
two variants that remove either relative-trajectory consistency or
absolute-position consistency.  These variants isolate the contribution of
each geometric component to predictive repair.

All repair variants share the initialization, training data, sampling budget,
and negative fine-tuning formulation; they differ in how generated futures
are selected and weighted.

The evaluation uses 500 randomized validation episodes from RoboTwin~2.0.  For
each episode, every video model generates one future conditioned on the
same observation and demonstrated action.  PSNR and SSIM compare the generated
future with its observed counterpart and measure visual fidelity.  The
action-following metrics below instead evaluate whether the generated robot
motion realizes the conditioning action.

\paragraph{Action-following metrics.}
We use the trajectory reliability indicator $C(y)$ defined in
Section~\ref{sec:predictive_repair}.  Because each episode provides one
generated future, we omit the sample index $k$ from $\hat y_{i,k}$ and
$\Delta R_{i,k}$.  Let
\begin{equation}
    \mathcal I_{\mathrm{ref}}
    =
    \{i:C(y_i^\star)=1\},
    \qquad
    N_{\mathrm{ref}}
    =
    |\mathcal I_{\mathrm{ref}}|,
    \label{eq:app_reliable_reference_set}
\end{equation}
denote the episodes whose observed future provides a reliable reference
trajectory.  The tracking success rate is the proportion of generated futures
that satisfy this reliability criterion:
\begin{equation}
    \mathrm{Tracking\ success\ rate}
    =
    \frac{1}{N_{\mathrm{ref}}}
    \sum_{i\in\mathcal I_{\mathrm{ref}}}
    C(\hat y_i).
    \label{eq:app_af_trackability}
\end{equation}
For a trackable generated future, the observed-reference gap is
\begin{equation}
    \Delta R_i
    =
    R_{\mathrm{AF}}^{\mathrm{obs}}(y_i^\star,a_i^\star)
    -
    R_{\mathrm{AF}}^{\mathrm{gen}}(\hat y_i,a_i^\star).
    \label{eq:app_af_evaluation_gap}
\end{equation}
We use $m_{\mathrm{eval}}=0.02$ for all reported diagnostic results.  The
action-following failure rate is
\begin{equation}
    \mathrm{AF\ failure\ rate}
    =
    \frac{1}{N_{\mathrm{ref}}}
    \sum_{i\in\mathcal I_{\mathrm{ref}}}
    \mathbf 1\!\left[
        C(\hat y_i)=0
        \;\lor\;
        \bigl(
        C(\hat y_i)=1
        \land
        \Delta R_i>m_{\mathrm{eval}}
        \bigr)
    \right].
    \label{eq:app_af_failure}
\end{equation}
We report both quantities as percentages.  Tracking success measures whether the
generated future contains sufficient robot motion for trajectory comparison,
whereas the failure rate additionally includes trackable futures whose
action-following score is substantially below the observed reference.

\begin{table}[t]
    \centering
    \caption{Evaluation and component analysis of predictive repair on 500
    randomized validation episodes from RoboTwin~2.0, with one generated
    future per episode.  Action-following metrics are evaluated on episodes
    whose observed future provides a reliable reference trajectory.}
    \label{tab:predictive_diagnostics}
    \small
    \setlength{\tabcolsep}{4pt}
    \begin{tabular}{lcccc}
        \toprule
        Predictive post-training
        & PSNR
        & SSIM
        & \shortstack{Tracking success\\(\%)}
        & \shortstack{AF failure\\(\%)} \\
        \midrule
        Continued flow matching
        & 30.69 & 0.9500 & 93.0 & 32.0 \\
        Repair with PSNR
        & 31.94 & 0.9576 & 96.8 & 18.6 \\
        Ours w/o relative-trajectory consistency
        & 32.25 & 0.9584 & 97.2 & 17.0 \\
        Ours w/o absolute-position consistency
        & 32.52 & 0.9607 & 98.0 & 10.6 \\
        Ours
        & \textbf{33.17} & \textbf{0.9642}
        & \textbf{99.2} & \textbf{6.8} \\
        \bottomrule
    \end{tabular}
\end{table}

\paragraph{Results.}
Repair with PSNR reduces the action-following failure rate from
32.0\% to 18.6\%, indicating that generic visual feedback corrects some
action-following errors.  The complete action-following score further reduces
the failure rate to 6.8\% and increases tracking success from 93.0\% to
99.2\%.  It also achieves the highest PSNR and SSIM.  These results show that
action-following feedback corrects prediction errors that are not captured by
visual similarity, without trading off standard video fidelity.

Among the individual geometric components, relative-trajectory consistency
provides the stronger action-following signal, yielding a failure rate of
10.6\%, compared with 17.0\% for absolute-position consistency.  Combining
both components further reduces the failure rate to 6.8\% and achieves the
highest trackability, PSNR, and SSIM.  The two terms are therefore
complementary: relative trajectories characterize the execution of the
conditioning action, while absolute positions retain information about
kinematic alignment.

\paragraph{Effect of the repair score on policy success.}
We use the PSNR-repaired video expert from
Table~\ref{tab:predictive_diagnostics} in the same action rejection
procedure and compare the resulting policy with \method{}.
As shown in Table~\ref{tab:repair_score_policy}, repair with the
kinematic alignment score improves success over PSNR-based repair
by 1.7 and 1.9 percentage points in clean and randomized
environments, respectively.
This comparison shows that the benefit of motion-based repair
extends beyond video metrics to policy success.

\begin{table}[t]
    \centering
    \caption{Policy success rates (\%) with different predictive
    repair scores and the same action rejection procedure on RoboTwin~2.0.}
    \label{tab:repair_score_policy}
    \small
    \begin{tabular}{lcc}
        \toprule
        Predictive repair score & Clean & Randomized \\
        \midrule
        PSNR & 92.1 & 91.8 \\
        Kinematic alignment (ours) & \textbf{93.8} & \textbf{93.7} \\
        \bottomrule
    \end{tabular}
\end{table}

\subsection{Action Negative Selection from Imagined Consequences}
\label{app:negative_selection}

\paragraph{Evaluation protocol.}
We compare continued flow matching and predictive repair in the clean
RoboTwin setting across all 50 tasks. For each task, 10 independently
generated successful scripted-expert trajectories provide one context each,
giving 500 contexts. The decision point is placed at 60\% of the valid
32-step action-chunk start range. The next 32 expert actions form the
reference continuation; completion of the entire task within this horizon
is not required.

For each context, four candidate action chunks are sampled once from a
fixed policy and reused for both video models. Each candidate and the expert
continuation are executed from the same restored simulator state, producing
2,000 candidate executions and 500 reference executions. Each video model
also predicts futures under the expert continuation and the identical
candidate actions. This diagnostic uses simulator executions to evaluate
the consequence comparisons, rather than as supervision for the
post-training objective.

\paragraph{Reference labels and selection metrics.}
The same frozen STEAM evaluator scores actual and imagined outcomes.
Let $y^{\mathrm{exec}}_{i,k}$ denote the observed consequence of executing
candidate $\hat a_{i,k}$, and let $y_i^\star$ denote the observed expert
continuation. The execution-based consequence gap is
\begin{equation}
    \Delta A^{\mathrm{exec}}_{i,k}
    = A_{\mathrm{STEAM}}(x_i,y_i^\star)
      - A_{\mathrm{STEAM}}(x_i,y^{\mathrm{exec}}_{i,k}).
    \label{eq:app_execution_gap}
\end{equation}
A candidate receives a reference negative label when
$\Delta A^{\mathrm{exec}}_{i,k}>m$.
For each video model, the imagined gap $\Delta A_{i,k}$ is computed as in
Equation~\ref{eq:relative_gap}, using that model for both the expert and
candidate futures. The candidate is selected when
$\Delta A_{i,k}>m$. We evaluate $m\in\{0.02,0.05,0.08\}$, using the
same threshold for reference labels and predicted selections within each
comparison; 0.08 is the margin used for action post-training in simulation.

We treat the reference negative label as the positive class for this
diagnostic. Thus, a true positive (TP) is a selected action that also receives
a negative label from its actual execution; a false positive (FP) is selected
only from its imagined consequence. False negatives (FN) and true negatives
(TN) are the corresponding unselected cases. We compute
\begin{equation}
    \mathrm{Precision}=\frac{\mathrm{TP}}{\mathrm{TP}+\mathrm{FP}},
    \qquad
    \mathrm{Recall}=\frac{\mathrm{TP}}{\mathrm{TP}+\mathrm{FN}},
    \qquad
    \mathrm{F1}=\frac{2\mathrm{TP}}
    {2\mathrm{TP}+\mathrm{FP}+\mathrm{FN}}.
    \label{eq:app_selection_metrics}
\end{equation}
All metrics use counts pooled over the 2,000 candidates, rather than averages
of task-level metrics. The reference labels contain 1,468, 1,129, and
755 negatives at margins of 0.02, 0.05, and 0.08, respectively.
They measure relative progress according to STEAM on actual executions,
not binary task success.

\begin{table}[t]
    \centering
    \caption{Action negative selection against reference labels from actual
    executions. Counts are pooled over 2,000 shared candidate actions;
    precision, recall, and F1 are percentages. The positive class is an
    action whose execution-based consequence gap exceeds the listed margin.}
    \label{tab:negative_selection_counts}
    \small
    \setlength{\tabcolsep}{3pt}
    \begin{tabular}{@{}llrrrrrrr@{}}
        \toprule
        Margin & Video model & TP & FP & FN & TN & Precision & Recall & F1 \\
        \midrule
        0.02 & Continued FM & 652 & 196 & 816 & 336 & 76.9 & 44.4 & 56.3 \\
             & Predictive repair & 935 & 239 & 533 & 293
             & \textbf{79.6} & \textbf{63.7} & \textbf{70.8} \\
        \midrule
        0.05 & Continued FM & 437 & 254 & 692 & 617 & 63.2 & 38.7 & 48.0 \\
             & Predictive repair & 640 & 193 & 489 & 678
             & \textbf{76.8} & \textbf{56.7} & \textbf{65.2} \\
        \midrule
        0.08 & Continued FM & 268 & 147 & 487 & 1,098 & 64.6 & 35.5 & 45.8 \\
             & Predictive repair & 382 & 165 & 373 & 1,080
             & \textbf{69.8} & \textbf{50.6} & \textbf{58.7} \\
        \bottomrule
    \end{tabular}
\end{table}

\FloatBarrier
\section{Real-World Evaluation Details}
\label{app:real_robot}

\subsection{Robot Platform and Policy Deployment}
\label{app:real_robot_setup}
\begin{figure}[t]
    \centering
    \includegraphics[width=0.8\linewidth]{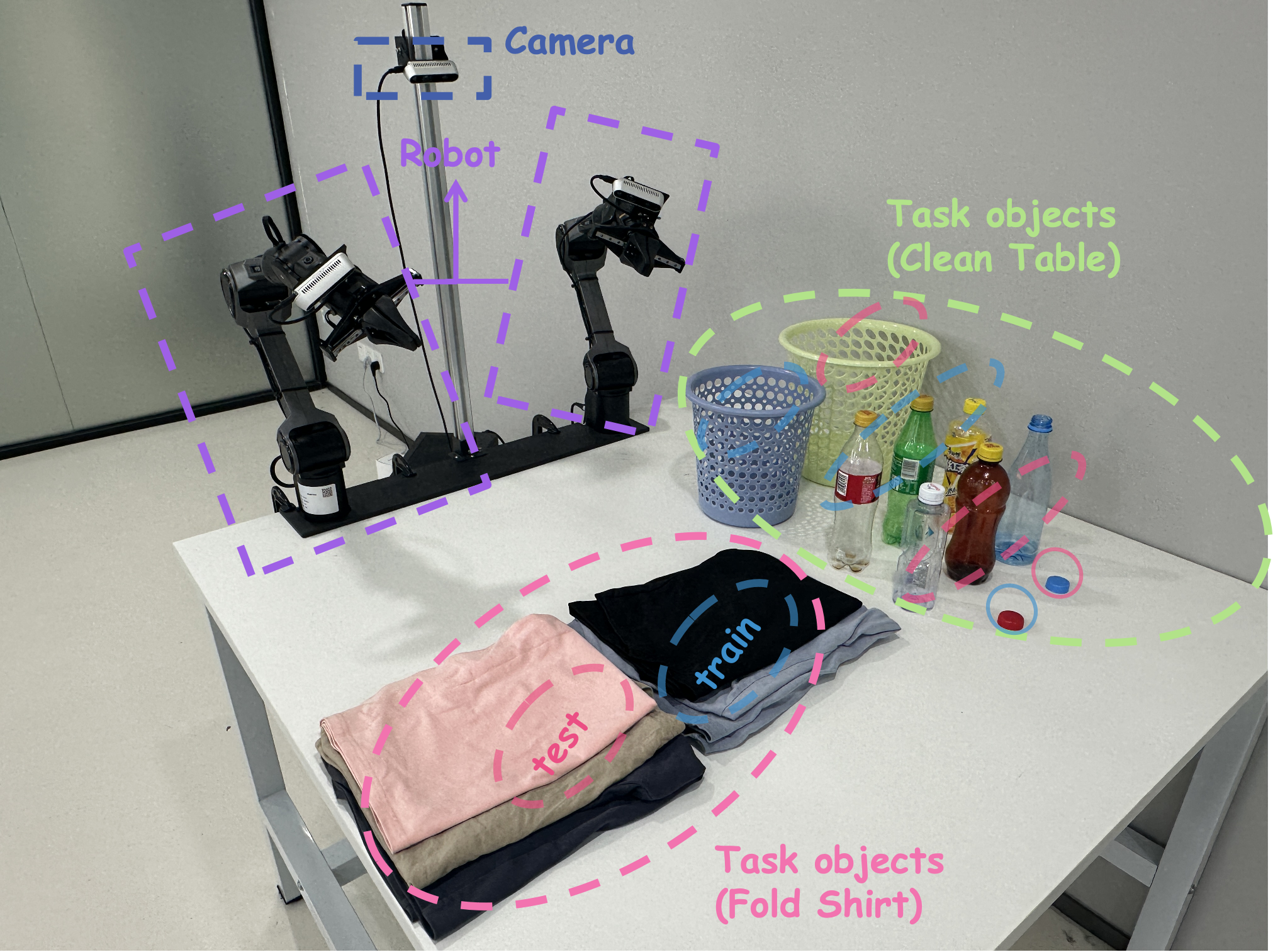}
    \caption{Real-world experimental setup, showing the dual-arm
    PIPER platform, cameras, and task objects for Fold Shirt
    and Clean Table.}
    \label{fig:real_robot_setup}
\end{figure}

We use a PIPER platform with two arms
(Figure~\ref{fig:real_robot_setup}). Each arm has six actuated joints and a
gripper with one degree of freedom.  Visual observations are provided by three RGB
cameras: one front camera and one camera mounted on each wrist.  The
raw camera streams are recorded at 480$\times$640 resolution and 30~Hz.  At
inference, the front view is resized to 320$\times$256, the two wrist views are
resized to 160$\times$128, and the views are assembled into a
384$\times$320 combined camera input (height $\times$ width).

The policy predicts a 20D absolute bimanual end-effector action,
comprising position, a 6D rotation representation, and gripper state for each
arm.  The policy uses 10 Euler steps to generate a 32-step action chunk at
30~Hz, with an execution horizon of 24 control steps.

\subsection{Offline Real-World Demonstrations}
\label{app:real_robot_data}

All task adaptation and post-training use offline teleoperated demonstrations;
the method does not collect additional environment interactions.  Demonstrations
for the two evaluated tasks comprise 3,031 trajectories and approximately
42.9 hours of robot experience.
Table~\ref{tab:real_robot_data} gives the composition for each task.  Training
uses one instruction describing the overall task for each trajectory and does not require
subtask labels for the long horizon tasks.

\begin{table}[h]
    \centering
    \caption{Offline demonstrations for the two evaluated real-world tasks.}
    \label{tab:real_robot_data}
    \small
    \begin{tabular}{lrrr}
        \toprule
        Task & Episodes & Frames & Duration (h) \\
        \midrule
        Fold Shirt  & 605   & 3,110,127 & 28.80 \\
        Clean Table & 2,426 & 1,526,405 & 14.13 \\
        \midrule
        Total       & 3,031 & 4,636,532 & 42.93 \\
        \bottomrule
    \end{tabular}
\end{table}

\subsection{Controlled Real-World Variations}
\label{app:real_robot_protocol}

\begin{figure}[!htbp]
    \centering
    \includegraphics[width=\linewidth]{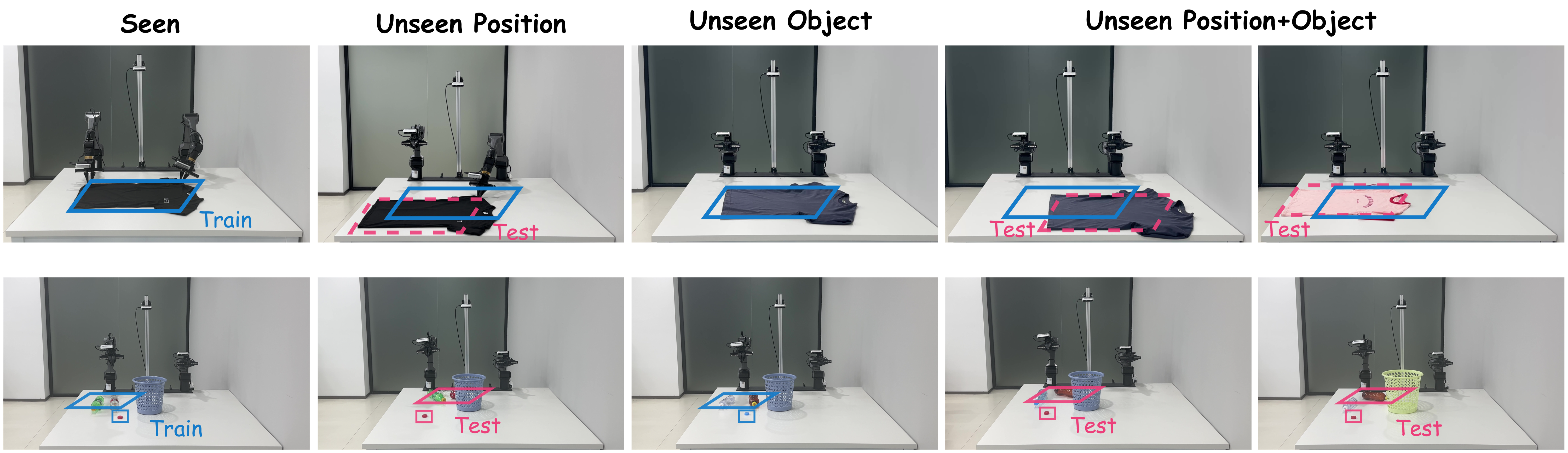}
    \caption{Representative evaluation configurations for Fold Shirt (top)
    and Clean Table (bottom). Blue and pink outlines indicate training
    and test position regions, respectively.}
    \label{fig:seen_unseen_configurations}
\end{figure}

We independently vary whether the initial position and object instance are
represented in the demonstrations, yielding four evaluation conditions.
Unseen conditions introduce position or object-instance variations
for one or more objects; the remaining objects may retain their
training configurations. Figure~\ref{fig:seen_unseen_configurations}
shows representative configurations for both tasks.
For every method and task, each condition
contains 20 autonomous episodes, for a total of 80.  Configuration lists are
fixed before policy evaluation and shared across all methods.

\paragraph{Seen.}
The objects and initial position region are represented in the demonstrations.
These trials quantify performance under familiar deployment conditions without
replaying demonstration trajectories.

\paragraph{Unseen Position.}
We retain familiar object instances while varying the initial
placement of one or more objects beyond their training regions.
For Fold Shirt, evaluation uses five test regions displaced from
the central training region.
For Clean Table, bottles and bottle caps may occupy either their
training or test regions within the same scene.

\paragraph{Unseen Object.}
We retain the training position regions while introducing one or
more object instances not used in training.
Fold Shirt uses two training shirts and three unseen shirts.
For Clean Table, the training objects comprise three bottles,
one bottle cap, and one trash bin; the unseen set contains three
additional bottles, one bottle cap, and one trash bin.

\paragraph{Unseen Position + Object.}
This condition combines both variations within the same scene,
including unseen object instances and initial placements outside
the training regions.

\FloatBarrier

\subsection{Success Criteria and Statistical Reporting}
\label{app:real_robot_statistics}

Each quantitative episode runs without human intervention and terminates upon
task completion, a time limit defined for each task, or a predefined failure or safety
condition.  Completion criteria are fixed before model evaluation and applied
identically to all methods.

Appendix Tables~\ref{tab:real_robot_factor_results} and
\ref{tab:clean_table_factor_results} report the success rate and
95\% Wilson score interval for every task and condition, each computed from 20
binary outcomes, corresponding to Figure~\ref{fig:real_robot_main}(a--b).
The average success under unseen conditions is the unweighted mean over
the two tasks and the three conditions with an unseen factor.

\begin{table}[!htbp]
    \centering
    \caption{Fold Shirt success rates (\%) for each real-world condition.
    Each condition contains 20 autonomous trials. Success rates appear above
    their 95\% Wilson score confidence intervals.}
    \label{tab:real_robot_factor_results}
    \small
    \setlength{\tabcolsep}{3pt}
    \renewcommand{\arraystretch}{1.2}
    \begin{tabular}{@{}lcccc@{}}
        \toprule
        Method & Seen & \shortstack{Unseen\\position}
        & \shortstack{Unseen\\object} & \shortstack{Unseen position\\+ object} \\
        \midrule
        $\pi_{0.5}$ & \shortstack{65.0\\{[43.3, 81.9]}} & \shortstack{35.0\\{[18.1, 56.7]}} & \shortstack{50.0\\{[29.9, 70.1]}} & \shortstack{35.0\\{[18.1, 56.7]}} \\
        X-VLA & \shortstack{40.0\\{[21.9, 61.3]}} & \shortstack{15.0\\{[5.2, 36.0]}} & \shortstack{20.0\\{[8.1, 41.6]}} & \shortstack{10.0\\{[2.8, 30.1]}} \\
        Motus & \shortstack{0.0\\{[0.0, 16.1]}} & \shortstack{0.0\\{[0.0, 16.1]}} & \shortstack{0.0\\{[0.0, 16.1]}} & \shortstack{0.0\\{[0.0, 16.1]}} \\
        Fast-WAM & \shortstack{0.0\\{[0.0, 16.1]}} & \shortstack{0.0\\{[0.0, 16.1]}} & \shortstack{0.0\\{[0.0, 16.1]}} & \shortstack{0.0\\{[0.0, 16.1]}} \\
        \method & \shortstack{\textbf{90.0}\\{[69.9, 97.2]}} & \shortstack{\textbf{90.0}\\{[69.9, 97.2]}} & \shortstack{\textbf{85.0}\\{[64.0, 94.8]}} & \shortstack{\textbf{85.0}\\{[64.0, 94.8]}} \\
        \bottomrule
    \end{tabular}
\end{table}

\begin{table}[!htbp]
    \centering
    \caption{Clean Table success rates (\%) for each real-world condition.
    Each condition contains 20 autonomous trials. Success rates appear above
    their 95\% Wilson score confidence intervals.}
    \label{tab:clean_table_factor_results}
    \small
    \setlength{\tabcolsep}{3pt}
    \renewcommand{\arraystretch}{1.2}
    \begin{tabular}{@{}lcccc@{}}
        \toprule
        Method & Seen & \shortstack{Unseen\\position}
        & \shortstack{Unseen\\object} & \shortstack{Unseen position\\+ object} \\
        \midrule
        $\pi_{0.5}$ & \shortstack{70.0\\{[48.1, 85.5]}} & \shortstack{45.0\\{[25.8, 65.8]}} & \shortstack{60.0\\{[38.7, 78.1]}} & \shortstack{40.0\\{[21.9, 61.3]}} \\
        X-VLA & \shortstack{50.0\\{[29.9, 70.1]}} & \shortstack{20.0\\{[8.1, 41.6]}} & \shortstack{40.0\\{[21.9, 61.3]}} & \shortstack{20.0\\{[8.1, 41.6]}} \\
        Motus & \shortstack{45.0\\{[25.8, 65.8]}} & \shortstack{35.0\\{[18.1, 56.7]}} & \shortstack{45.0\\{[25.8, 65.8]}} & \shortstack{20.0\\{[8.1, 41.6]}} \\
        Fast-WAM & \shortstack{55.0\\{[34.2, 74.2]}} & \shortstack{40.0\\{[21.9, 61.3]}} & \shortstack{35.0\\{[18.1, 56.7]}} & \shortstack{25.0\\{[11.2, 46.9]}} \\
        \method & \shortstack{\textbf{85.0}\\{[64.0, 94.8]}} & \shortstack{\textbf{75.0}\\{[53.1, 88.8]}} & \shortstack{\textbf{75.0}\\{[53.1, 88.8]}} & \shortstack{\textbf{70.0}\\{[48.1, 85.5]}} \\
        \bottomrule
    \end{tabular}
\end{table}

\FloatBarrier

\subsection{Component Ablations on Fold Shirt}
\label{app:real_robot_ablation}

We evaluate predictive repair and action rejection separately on Fold Shirt
(Table~\ref{tab:fold_shirt_ablation}). Each variant is evaluated over
20 autonomous trials in each of the four conditions described in
Appendix~\ref{app:real_robot_protocol}, totaling 80 trials.
Repair / GT FM applies predictive
repair to the video expert while training the action expert with
demonstration flow matching. GT FM / Rejection trains the video expert
with demonstration flow matching and uses its imagined consequences for
action rejection. Neither single-stage variant completes the task in these trials.
Predictive repair alone does not yield successful task completion
when the action expert is trained solely with demonstration flow
matching. Conversely, action rejection without predictive repair
also remains insufficient.
Combining the two stages achieves 87.5\% average success, supporting
their complementary roles in long-horizon shirt folding:
improving the action fidelity of predicted consequences and using
that feedback to refine sampled actions.

\begin{table}[!htbp]
    \centering
    \caption{Component ablation on Fold Shirt. Success rates are averaged
    over the four evaluation conditions. GT FM denotes flow matching
    on demonstrations.}
    \label{tab:fold_shirt_ablation}
    \small
    \setlength{\tabcolsep}{8pt}
    \begin{tabular}{llc}
        \toprule
        Video & Action & Average success (\%) \\
        \midrule
        Repair & GT FM & 0.0 \\
        GT FM & Rejection & 0.0 \\
        Repair & Rejection & \textbf{87.5} \\
        \bottomrule
    \end{tabular}
\end{table}

\subsection{Real-World Execution Examples}
\begin{figure}[!htbp]
    \centering
    \includegraphics[
        width=\textwidth,
        height=0.85\textheight,
        keepaspectratio
    ]{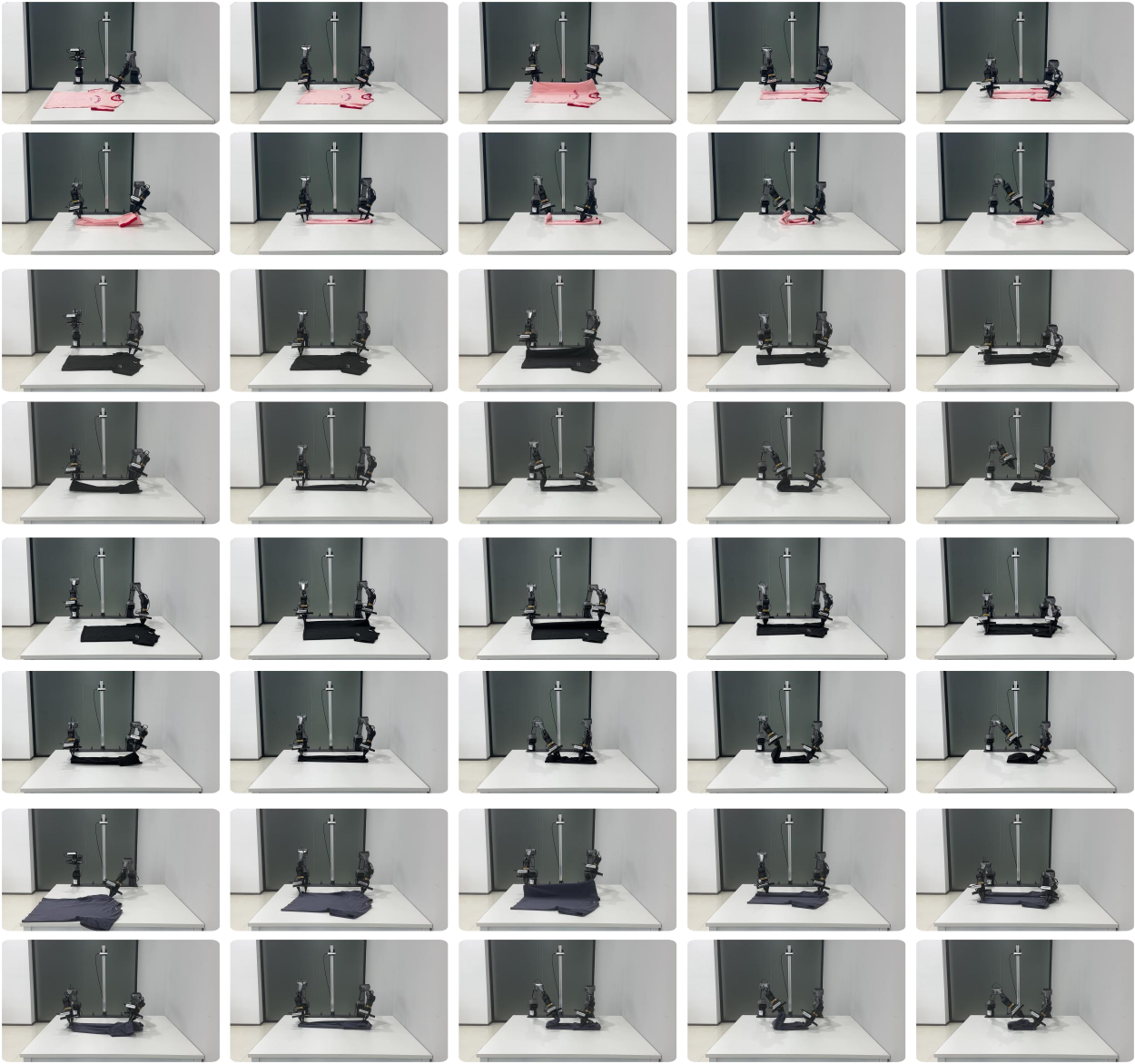}
    \caption{Execution sequences of \method{} on Fold Shirt.
    Each sequence spans two consecutive rows and proceeds
    from left to right.}
    \label{fig:real_demo1}
\end{figure}

\begin{figure}[p]
    \centering
    \includegraphics[
        width=\textwidth,
        height=0.85\textheight,
        keepaspectratio
    ]{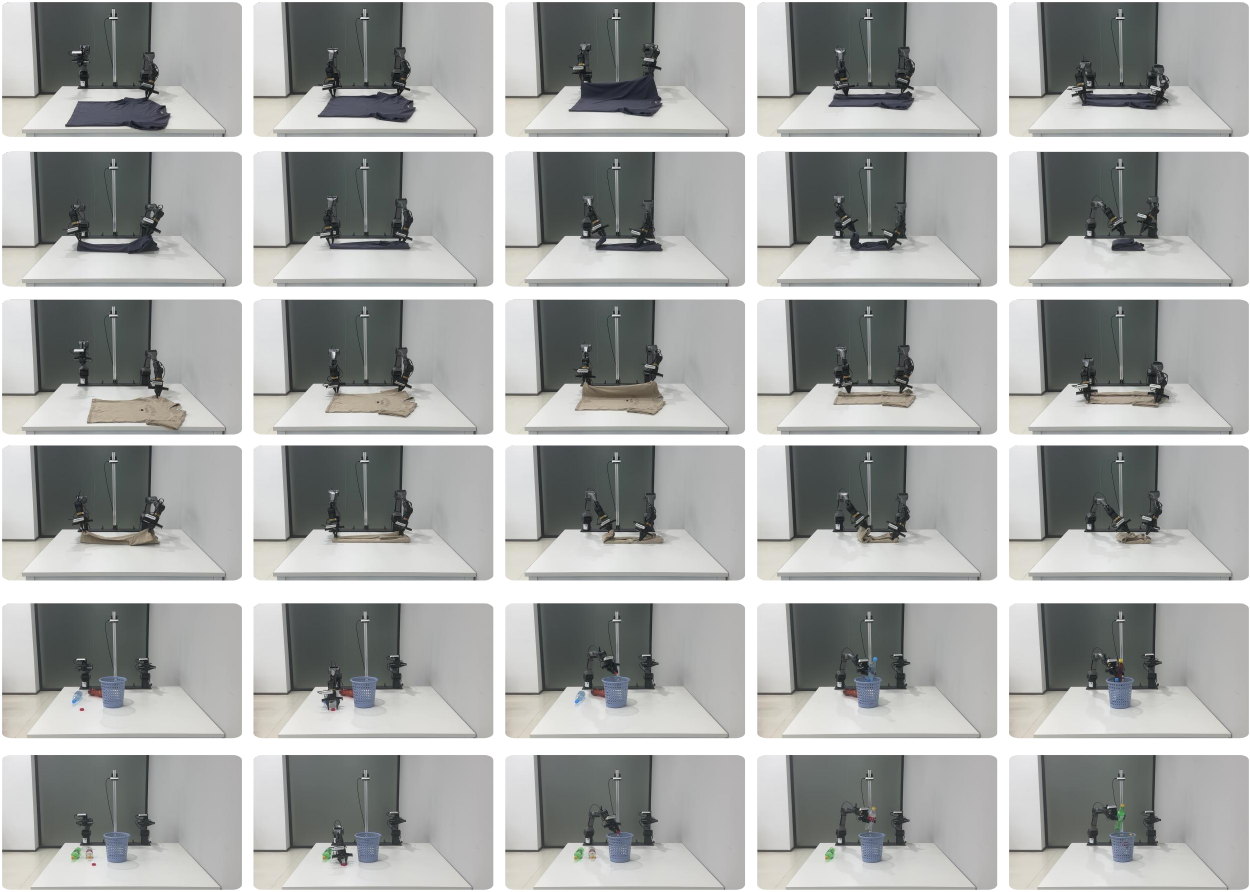}
    \caption{Execution sequences of \method{} on Fold Shirt
    (top four rows) and Clean Table (bottom two rows).
    Each Fold Shirt sequence spans two rows, while each Clean Table
    sequence occupies one row. Frames proceed from left to right.}
    \label{fig:real_demo2}
\end{figure}

Figures~\ref{fig:real_demo1} and~\ref{fig:real_demo2}
show representative execution sequences of \method{} on
Fold Shirt and Clean Table.

